\documentclass[conference]{IEEEtran}
\IEEEoverridecommandlockouts
\usepackage{cite}
\usepackage{amsmath,amssymb,amsfonts}
\usepackage{algorithm}
\usepackage{algpseudocode}
\usepackage{graphicx}
\usepackage{subcaption}
\usepackage{standalone}
\usepackage{hyperref}
\usepackage{textcomp}
\usepackage{xcolor}

\definecolor{darkWhite}{rgb}{0.96,0.96,0.96}
\definecolor{bluekeywords}{rgb}{0.13,0.13,1}
\definecolor{greencomments}{rgb}{0,0.5,0}
\definecolor{redstrings}{rgb}{0.9,0,0}

\definecolor{Comment}{RGB}{97,161,176}

\definecolor{btfGreen}{RGB}{51,160,44}
\definecolor{btfRed}{RGB}{190,60,90}

\definecolor{bleuUni}{RGB}{0, 157, 224}
\definecolor{marronUni}{RGB}{68, 58, 49}

\definecolor{bluecite}{HTML}{009DE0}

\definecolor{Paired_1}{RGB}{31,120,180}
\definecolor{Paired_2}{RGB}{166,206,227}
\definecolor{Paired_3}{RGB}{51,160,44}
\definecolor{Paired_4}{RGB}{178,223,138}
\definecolor{Paired_5}{RGB}{227,26,28}
\definecolor{Paired_6}{RGB}{251,154,153}
\definecolor{Paired_7}{RGB}{255,127,0}
\definecolor{Paired_8}{RGB}{253,191,111}
\definecolor{Paired_9}{RGB}{106,61,154}
\definecolor{Paired_10}{RGB}{202,178,214}
\definecolor{Paired_11}{RGB}{177,89,40}
\definecolor{Paired_12}{RGB}{255,255,153}
\definecolor{Accent_1}{RGB}{127,201,127}
\definecolor{Accent_2}{RGB}{190,174,212}
\definecolor{Accent_3}{RGB}{253,192,134}
\definecolor{Accent_4}{RGB}{255,255,153}
\definecolor{Accent_5}{RGB}{56,108,176}
\definecolor{Accent_6}{RGB}{240,2,127}
\definecolor{Accent_7}{RGB}{191,91,23}
\definecolor{Accent_8}{RGB}{102,102,102}
\definecolor{Spectral_1}{RGB}{158,1,66}
\definecolor{Spectral_2}{RGB}{213,62,79}
\definecolor{Spectral_3}{RGB}{244,109,67}
\definecolor{Spectral_4}{RGB}{253,174,97}
\definecolor{Spectral_5}{RGB}{254,224,139}
\definecolor{Spectral_6}{RGB}{255,255,191}
\definecolor{Spectral_7}{RGB}{230,245,152}
\definecolor{Spectral_8}{RGB}{171,221,164}
\definecolor{Spectral_9}{RGB}{102,194,165}
\definecolor{Spectral_10}{RGB}{50,136,189}
\definecolor{Spectral_11}{RGB}{94,79,162}
\definecolor{Set1_1}{RGB}{228,26,28}
\definecolor{Set1_2}{RGB}{55,126,184}
\definecolor{Set1_3}{RGB}{77,175,74}
\definecolor{Set1_4}{RGB}{152,78,163}
\definecolor{Set1_5}{RGB}{255,127,0}
\definecolor{Set1_6}{RGB}{255,255,51}
\definecolor{Set1_7}{RGB}{166,86,40}
\definecolor{Set1_8}{RGB}{247,129,191}
\definecolor{Set1_9}{RGB}{153,153,153}
\definecolor{Set1_10}{RGB}{0,0,0}
\definecolor{Set2_1}{RGB}{102,194,165}
\definecolor{Set2_2}{RGB}{252,141,98}
\definecolor{Set2_3}{RGB}{141,160,203}
\definecolor{Set2_4}{RGB}{231,138,195}
\definecolor{Set2_5}{RGB}{166,216,84}
\definecolor{Set2_6}{RGB}{255,217,47}
\definecolor{Set2_7}{RGB}{229,196,148}
\definecolor{Set2_8}{RGB}{179,179,179}
\definecolor{Dark2_1}{RGB}{27,158,119}
\definecolor{Dark2_2}{RGB}{217,95,2}
\definecolor{Dark2_3}{RGB}{117,112,179}
\definecolor{Dark2_4}{RGB}{231,41,138}
\definecolor{Dark2_5}{RGB}{102,166,30}
\definecolor{Dark2_6}{RGB}{230,171,2}
\definecolor{Dark2_7}{RGB}{166,118,29}
\definecolor{Dark2_8}{RGB}{102,102,102}
\definecolor{Reds_1}{RGB}{255,245,240}
\definecolor{Reds_2}{RGB}{254,224,210}
\definecolor{Reds_3}{RGB}{252,187,161}
\definecolor{Reds_4}{RGB}{252,146,114}
\definecolor{Reds_5}{RGB}{251,106,74}
\definecolor{Reds_6}{RGB}{239,59,44}
\definecolor{Reds_7}{RGB}{203,24,29}
\definecolor{Reds_8}{RGB}{165,15,21}
\definecolor{Reds_9}{RGB}{103,0,13}
\definecolor{Greens_1}{RGB}{247,252,245}
\definecolor{Greens_2}{RGB}{229,245,224}
\definecolor{Greens_3}{RGB}{199,233,192}
\definecolor{Greens_4}{RGB}{161,217,155}
\definecolor{Greens_5}{RGB}{116,196,118}
\definecolor{Greens_6}{RGB}{65,171,93}
\definecolor{Greens_7}{RGB}{35,139,69}
\definecolor{Greens_8}{RGB}{0,109,44}
\definecolor{Greens_9}{RGB}{0,68,27}
\definecolor{Blues_1}{RGB}{247,251,255}
\definecolor{Blues_2}{RGB}{222,235,247}
\definecolor{Blues_3}{RGB}{198,219,239}
\definecolor{Blues_4}{RGB}{158,202,225}
\definecolor{Blues_5}{RGB}{107,174,214}
\definecolor{Blues_6}{RGB}{66,146,198}
\definecolor{Blues_7}{RGB}{33,113,181}
\definecolor{Blues_8}{RGB}{8,81,156}
\definecolor{Blues_9}{RGB}{8,48,107}

\def\BibTeX{{\rm B\kern-.05em{\sc i\kern-.025em b}\kern-.08em
    T\kern-.1667em\lower.7ex\hbox{E}\kern-.125emX}}
\usepackage{tikz}
\usepackage{pgfplots}
\pgfplotsset{compat=newest}
\usetikzlibrary{er, positioning, calc}

\usepackage{multirow}

\begin{document}


\title{FLoCoRA: \MakeUppercase{Federated Learning Compression with Low-Rank Adaptation}\\
\thanks{This work is supported by the \textit{Futur et Ruptures} program funded by IMT and Institut Carnot TSN, and by the GdR IASIS.}}



\author{
    \IEEEauthorblockN{
    Lucas Grativol\IEEEauthorrefmark{1,2},
    Mathieu Léonardon\IEEEauthorrefmark{1},
    Guillaume Muller\IEEEauthorrefmark{3}, 
    Virginie Fresse\IEEEauthorrefmark{2} and Matthieu Arzel\IEEEauthorrefmark{1}}
    
    \IEEEauthorblockA{\IEEEauthorrefmark{1}IMT Atlantique, Lab-STICC, UMR CNRS 6285, F-29238 Brest, France}
    \IEEEauthorblockA{\IEEEauthorrefmark{2}Hubert Curien Laboratory, Saint-Etienne, France}
    \IEEEauthorblockA{\IEEEauthorrefmark{3}Mines Saint-Etienne, Institut Henri Fayol, Saint-Etienne, France}
}

\maketitle

\begin{abstract}
Low-Rank Adaptation (LoRA) methods have gained popularity in efficient parameter fine-tuning of models containing hundreds of billions of parameters. In this work, instead, we demonstrate the application of LoRA methods to train small-vision models in Federated Learning (FL) from scratch. We first propose an aggregation-agnostic method to integrate LoRA within FL, named FLoCoRA, showing that the method is capable of reducing communication costs by 4.8 times, while having less than 1\% accuracy degradation, for a CIFAR-10 classification task with a ResNet-8. Next, we show that the same method can be extended with an affine quantization scheme, dividing the communication cost by 18.6 times, while comparing it with the standard method, with still less than 1\% of accuracy loss, tested with on a ResNet-18 model. Our formulation represents a strong baseline for message size reduction, even when compared to conventional model compression works, while also reducing the training memory requirements due to the low-rank adaptation.
\end{abstract}


\section{Introduction}

The preservation of data privacy has long been a goal in the development of approaches to train machine learning models. In traditional machine learning, raw data from embedded systems are sent over a network to a powerful server for model training, raising concerns about confidentiality~\cite{sun2014data}. Federated Learning (FL) has emerged as a promising mitigation of these problems~\cite{mcmahan2017communication}. In FL, each participating client keeps its data locally, only sharing the results of its local training. These results, often in the form of model parameters or gradients, are fused in a central orchestration server. As the training process shifts from a server to a third-party device, there is a trade-off between increased privacy and computational and communication overhead. The challenge of convergence in FL~\cite{kairouz2021advances} further supports this idea, as the diverse data distribution among devices leads to conflicting update models, which affects the convergence time.

To overcome the computation and communication problem in FL, previous works have focused on model compression techniques such as pruning~\cite{grativol2023federated}, quantization~\cite{reisizadeh2020fedpaq} and more recently the application of low-rank adaptation (LoRA) with~\cite{hyeon2021fedpara,babakniya2023slora,cho2023heterogeneous}, following the original work on LoRA~\cite{hu2021lora}. Unlike previous works, that focused on large models, we demonstrate the application of LoRA for training small-vision models from scratch. Going further, we demonstrate how quantization can be applied to LoRA and FL, dividing the cost of communication by 18.6, compared to the standard method, while keeping the accuracy degradation within 1\%.
\\\\
We can summarize our contributions as : 

\begin{itemize}
  \item  We demonstrate how LoRA adapters can be integrated to the FL framework, while still being implementable in any FL optimization method, proposing FLoCoRA. Representing a strong baseline for communication and memory reduction in FL. Our code is publicly accessible~\footnote{\url{https://anonymous.4open.science/r/FLoCoRA_eusipco24-A6F1/}}.
  \item We study the impact of LoRA hyperparameters for a classification task with small vision models, which allows these models to be trained from scratch, contributing to reduce message sizes in FL up to 4.8 times, with 1\% loss of accuracy for a ResNet-8.
  \item We introduce an affine quantization scheme with FLoCoRA, allowing further compression rates of 18.6 to 37.3 times with up to 1\% loss of accuracy for a ResNet-18.
\end{itemize}
 
\section{Background}
\subsection{Federated Learning}
\label{sec:federated_learning}

As the most used aggregation method in FL, we use FedAvg (Federated Averaging)~\cite{mcmahan2017communication} as the showcase for a typical FL framework, while also presenting the baseline used in this work. It operates by sampling a subset $K$, of a pool of clients $C$, in each round to train a model with parameters $w$ for a specified number of local epochs. Each client has its own dataset of size $n_i$, and the total size of the datasets of the participating clients in a round is denoted $n$. Each client $k \in K$, seeks to find the parameters $w$ that minimize its local loss $f_k(w) = \mathbb{E}[l(X,y,w)]$, for a set of examples $X$ and labels $y$, and the loss function $l(.)$. The general objective of FedAvg is then to find a global $w$ that minimizes Equation~\ref{eq1}, where $F_k(w) = \mathbb{E}[f_k]$. When this process is iterating for a certain number of rounds $R$, the final result is expected to increase the individual performance of each client~\cite{tan2022towards}, without the need for data sharing.

\begin{equation}
\begin{aligned}
\label{eq1}
\underset{w}{\mathrm{min}}~f(w) = \sum\limits_{k=1}^{K} \frac{n_i}{n} F_{k}(w)
 \end{aligned}
\end{equation}

During training, model parameters are downloaded from server to client and uploaded from client to server. Equation~\ref{eq2} represents the total communication cost (TCC) for a certain number of rounds $R$. Here, we consider $Q_p$ as the number of bits of each element in $w$ and $|w|$ as the total number of elements/parameters. To decrease the size of a message, one can reduce the number of parameters or reduce the value of $Q_p$. Equation~\ref{eq2} can be utilized to calculate the communication cost for a single client in FL. Therefore, we employ this equation throughout this work to determine the TCC.

\begin{equation}
\begin{aligned}
\label{eq2}
TCC(R) = 2RQ_p|w|
 \end{aligned}
\end{equation}

\subsection{Model Compression}
\label{sec:msg_comp}

This is the point where traditional model compression techniques help the FL framework. Model compression is a widely adopted solution~\cite{cheng2017survey} to reduce computational and memory requirements in centralised machine learning. As such, the ideas of pruning~\cite{qiu2022zerofl} and quantization~\cite{reisizadeh2020fedpaq}, two widely adopted techniques in model compression, have been studied in the FL scenario~\cite{kairouz2021advances}.

Within the FL framework, pruning can be applied to induce more sparsity in a model, leading to more compressible parameters~\cite{grativol2023federated}, or to eliminate certain architectural elements, such as kernel or filters, to transmit fewer parameters~\cite{jiang2022model}. By introducing sparsity or removing certain parts of a model, pruning can also reduce the computation needs of a model during training~\cite{tessier2022rethinking}.

Another method, sometimes complementary to pruning~\cite{bai2023unified}, is quantization, where the objective is to reduce the binary representation of the parameters. As usually models are represented using 32- or 16-bits floating point (FP) numbers, the idea is to reduce it to 8- or sub-8-bit FP, or even change the data to fixed-point or integer formats~\cite{wu2018training}. 

Pruning has been used successfully in FL, either post-training to introduce sparsity and reduce message size, or during training to reduce computations and message size~\cite{qiu2022zerofl}. However, such methods require additional steps to induce sparsity, resulting in overhead.

\subsection{Low-Rank Adaptation}
\label{LoRA}

 In the domain of transfer learning, low-rank adaptation methods have gained popularity for task adaptation of model containing hundreds of billions of parameters~\cite{hu2021lora,bensaid2024novel}. LoRA comes from the need to adapt a large-scale already trained model to a new task, in order to save computation time and reduce the memory footprint. For example, the parameters , $W_l \in \mathbb{R}^{d_1 \times d_2}$, of pretrained linear layer, $l$, are reformulated with a new adapter, a parallel layer, obtaining $W^{*}_l=W_l+ \frac{\alpha}{r}BA$, where $A \in \mathbb{R}^{d_1 \times r}$ and $B \in \mathbb{R}^{r \times d_2}$ are two matrices of maximum rank $r$. The term $\frac{\alpha}{r}$ is a scaling factor, based on a hyperparameter $\alpha$ and rank $r$. The idea is to keep $W_l$ frozen and to train $BA$ to represent a lower-rank update of $W_l$. An advantage of this approach is that the matrix $BA$ can be incorporated back into the original pretrained weights, $W_l^*$, without any additional latency. LoRA has shown that for $r << min(d_1,d_2)$, $W_l$ can be adapted to a new task, largely reducing the need to fine-tune the entire model. 

When LoRA is applied to FL, the original model is kept frozen and we only need to train the adapters, $B$ and $A$. The adapters have fewer parameters than the original model, resulting in a lighter model for training because they require less memory for the gradients. Since the original model remains frozen, it is not necessary to communicate it between the server and client in every round, thus the communication is also lighter. 

Previous studies like SLoRA~\cite{babakniya2023slora} have investigated the potential impact of LoRA techniques on FL. They suggested merging standard FL training with matrix decomposition to achieve favorable initialization for both the original model and matrices $A$ and $B$. HLoRA~\cite{cho2023heterogeneous} allowed clients to select different LoRA ranks, according to their individual constraints, to adapt a pretrained base model. FedPara~\cite{hyeon2021fedpara}, redefined the LoRA adapter as $W^{*}_l = W_l+ (X_1.Y^T_1)\odot(X_2.Y^T_2)$, where $\odot$ is the Hadamard product, in order to obtain an update of higher rank than conventional LoRA. 

However, SLoRA and HLoRA have not investigated the application of LoRA methods for the training of small CNNs from scratch, having focused on how LoRA was applied to bigger models like large-language models (LLM) and foundation models. Closer to our work, FedPara~\cite{hyeon2021fedpara} has tested on small CNNs, but its low-rank method and quantization is limited to FP-16 and the adaptation of FedPaq~\cite{reisizadeh2020fedpaq} on a rather simple test scenario.

\section{Our method - FLoCoRA}
\label{sec:meth}

In this work, we propose never to update the parameters of the randomly initialized neural network parameters shared between clients at the start of training. Only adapters parameters will be trained, exchanged, and updated. Since the original parameters remain unchanged, the exchange of LoRA parameters is sufficient to represent the updates from each client. The server continues to receive updated parameters from clients, which means that this method can also be integrated with other FL techniques, without further changes.

\begin{figure}[ht]
    \centerline{
    \includegraphics[width=0.45\textwidth]{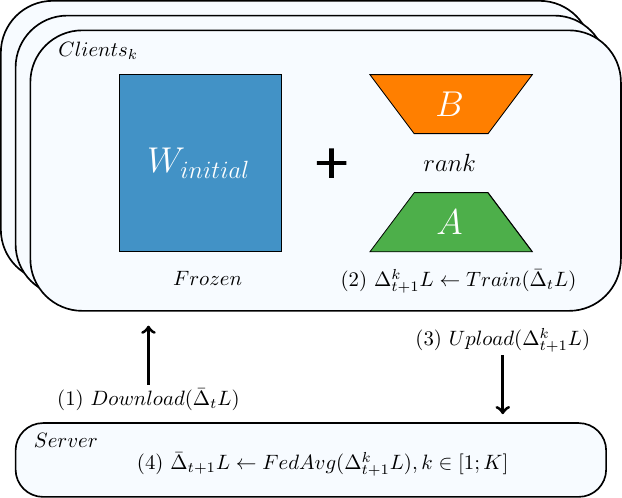}}
    \caption{FLoCoRA training loop.}
    \label{fedavglora}
\end{figure}


Figure~\ref{fedavglora} illustrates our approach, \textbf{F}ederated \textbf{L}earning \textbf{Co}mpression with \textbf{Lo}w-\textbf{R}ank \textbf{A}daptation (FLoCoRA), for a single communication round. All clients start with identical weights, denoted as $W_{initial}$, which remain fixed throughout the training process. In the first step $(1)$, the server transmits the global LoRA adapters parameters, $\bar\Delta_{t} L$, to the selected subset $K$ of clients. Subsequently, in step $(2)$, each client independently trains its LoRA adapter locally and uploads the result, $\Delta^k_{t+1} L$, to the server in step $(3)$. Finally, in step $(4)$, the server uses the same weighted averaging mechanism as FedAvg to obtain the updated global LoRA adapter parameters for the next round, denoted $\bar\Delta_{t+1} L$.

Table~\ref{tab:rsize} presents the size of the trainable parameters for a ResNet-8~\cite{he2016deep}, for different values of $r$. It should be noted that we adapt the convolution layers using LoRA, but the normalization and fully connected (FC) layers of the original model are trained; more details of this are discussed in Section~\ref{sec:experiments}.


\begin{table}[t]
\centering
\caption{Number of parameters for different sizes of r. For each value, we have the total number of parameters to be trained/sent and the total number of the parameters with the original model plus the LoRA adapter.}
\label{tab:rsize}
\begin{tabular}{lccc}
\hline
\multicolumn{1}{l}{\textbf{Method}} & \multicolumn{1}{l}{\textbf{\begin{tabular}[c]{@{}c@{}}Total \\ Params\end{tabular}}} & \multicolumn{1}{c}{\textbf{\begin{tabular}[c]{@{}c@{}}Trained\\ Params\end{tabular}}} & \textbf{\begin{tabular}[c]{@{}c@{}}\% of Trained\\ Params\end{tabular}} \\ \hline
\textbf{FedAvg} & \textbf{1.23M} & \textbf{1.23M} & \textbf{100} \\ \hline
FLoCoRA ($r=8$) & 1.30M & 69.45K & 5.35 \\
FLoCoRA ($r=16$) & 1.36M & 131.92K & 9.70 \\
FLoCoRA ($r=32$) & 1.48M & 256.84K & 17.30 \\
FLoCoRA ($r=64$) & 1.73M & 506.70K & 29.22 \\
FLoCoRA ($r=128$) & 2.23M & 1.00 M & 45.05 \\ \hline
\end{tabular}
\end{table}

For convolution layers, we follow the decomposition proposed in~\cite{huh2022low}. Let $P_l \in \mathbb{R}^{O \times I \times K_r \times  K_r}$ be a convolution layer; then we define its LoRA adapter matrices as $B \in \mathbb{R}^{r \times I \times K_r \times  K_r}$ and $A \in \mathbb{R}^{O \times r \times 1 \times  1}$. Output channels are denoted as $O$, the input channels as $I$, and the kernel size as $K_r$.

To verify the impacts of using LoRA with FL, we first investigate which layers of the original model need to be trained in a normal way and which can receive the LoRA adapters. Then, we compare our work with recent literature on compression methods for FL, showing that LoRA methods are a strong baseline for future works in the domain. Finally, we show that quantizing LoRA parameters can lead to additional message compression, even when dealing with highly diverse data distributions across clients.

\section{Experiments and Discussion}
\label{sec:experiments}

Our basic setup for FL consists of 100 clients, where at each round 10 \% of clients are sampled. We train for a total of 100 rounds. For simplicity, based on~\cite{hyeon2021fedpara}, we fix the batch size, the learning rate, the number of local epochs and momentum at 32, 0.01, 5 and 0.9, respectively, for all clients. We use SGD with momentum as the local optimizer for clients. We use FedAvg as our aggregation algorithm. We ran all the experiments three times with different seeds.

Initially, to verify how FLoCoRA training impacts convergence in FL, we trained a ResNet-8 for an image classification task with the CIFAR-10 dataset. Following an LDA distribution with parameter 0.5~\cite{mcmahan2017communication}, we also replace the batch normalization layer with the group normalization layer as suggested by~\cite{hsu2019measuring}. Table~\ref{tab:lora_ablation} presents an ablation of the layers to be trained with LoRA hyperparameters $r=32$ and $\alpha=512$. We chose to use $r=32$ because it was the minimum rank of which the accuracy had less than 1$\%$ of degradation. Starting with a randomly initialized ResNet-8 model, we freeze the entire model, introducing LoRA adapters for all convolutions and to the final FC layer, corresponding to "FLoCoRA Vanilla" in Table~\ref{tab:lora_ablation}. Subsequently, we unfreeze the normalization layers, "+ Norm. Layers", then we remove the LoRA adapter from the final FC layer and unfreeze it, "+ Final FC". 

\begin{table}[]
\centering
\caption{The effect of training different layers with or without LoRA adapters.}
\label{tab:lora_ablation}
\begin{tabular}{rcc}
\hline
\multicolumn{1}{c}{\textbf{Method}} & \textbf{\begin{tabular}[c]{@{}c@{}}Nb. of Params. \\ to update\end{tabular}} & \textbf{Accuracy} \\ \hline
\multicolumn{1}{c}{\textbf{FedAvg}} & \textbf{1.23 M} & \textbf{76.14} $\pm$ \textbf{0.74}                     \\ \hline
\multicolumn{1}{l}{FLoCoRA Vanilla}         & 0.26 M   & \multicolumn{1}{l}{22.14 $\pm$ 3.99} \\
+ Norm. layers                               & 0.26 M   & \multicolumn{1}{l}{39.80 $\pm$ 12.05} \\
+ Final FC                                   & 0.26 M   & \textbf{75.51 $\pm$ 1.34}         \\
\hline
\end{tabular}
\end{table}

Normalization layers need to be trained to capture running statistics, as it cannot be adapted with LoRA methods, so we need to train it. For the FC layer, we hypothesize that the lower rank is not enough for the last classification layer, as it is a layer that tends to be highly specialized and sensible~\cite{neyshabur2020being}. For the remainder of this paper, we keep this configuration for FLoCoRA experiments.

\begin{figure}[b]
    \centerline{\includegraphics[width=0.42\textwidth]{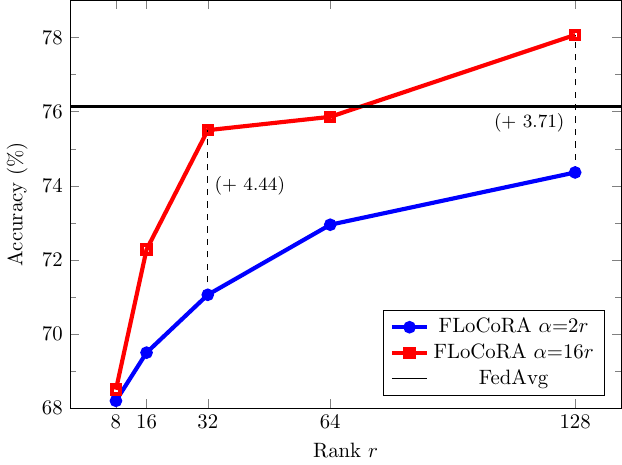}}
    \caption{The relationship between the $r$ hyperparameter in FLoCoRA and the scaling factor $\alpha$, in $\frac{\alpha}{r}$. Two scenarios are evaluated, $\alpha=2r$ and $\alpha=16r$, against FedAvg.}
    \label{fig:lora_r_alpha}
\end{figure}

Expanding on the importance of choosing the appropriate layers, we examine the trade-off between rank $r$ and the scaling parameter $\alpha$. We compare a FedAvg baseline with two different setups, where $\alpha$ is set to $2r$ and $16r$, the results are shown in Figure~\ref{fig:lora_r_alpha}. In LoRA's original paper~\cite{hu2021lora}, $\alpha$ is scaled to twice the rank for LLMs. However, when training small CNNs with FL, we found that increasing this factor further led to improvements in accuracy of up to 4.4\%. Increasing the scaling factor of the LoRA adapters layers is a way to adjust their learning rate to higher values while keeping more sensible layers, the normalization and the FC layers, in a lower learning rate, improving training stability and performance.

Looking at the results in Figure~\ref{fig:lora_r_alpha}, FLoCoRA is able to achieve an accuracy drop of less than 1$\%$, while sharing only 0.26M parameters, representing a reduction of 4.8$\times$ compared to sharing the entire model. For a rank of 128, the number of parameters to send is closer to the original model, as seen in Table~\ref{tab:rsize}, but we see an improvement in accuracy of $2\%$. Indeed, for the ResNet-8 design used, the convolution layers throughout the model have output channel numbers of 64, 128, or 256. Consequently, opting for a rank of 128 results in an augmentation of rank for certain layers while maintaining the same rank for others. This modification results in a greater overall rank update for the shallower convolution layers compared to the original model, and the low-rank update is actually applied to the deeper convolution layers. Finally, the model has slightly less parameters because the larger convolution layers, with 256 output channels, are adapted with a lower rank of 128, effectively reducing the number of parameters.

\begin{table}[b]
\centering
\caption{Total communication cost (TTC) for different quantization levels with FLoCoRA, for a rank of 32 and alpha of 512, during 100 rounds of FL.}
\label{tab:lora_quant}
\begin{tabular}{cclc}
\hline
\multicolumn{1}{c}{\textbf{Method}} &
  \multicolumn{1}{c}{\textbf{Quantization}} &
  \multicolumn{1}{c}{\textbf{\begin{tabular}[c]{@{}c@{}} TCC \end{tabular}}} &
  \textbf{Accuracy} \\ \hline
FedAvg              & FP   & 982.07 MB ($\div1$) & 76.14 $\pm$ 0.74     \\ \hline
FLoCoRA                 & FP   & 205.47 MB ($\div4.8$)  & 75.51 $\pm$ 1.34     \\
                     & int8 & 55.56 MB ($\div17.7$)  & 74.21 $\pm$ 1.05     \\
                     & int4 & 30.15 MB ($\div32.6$)  & 73.15 $\pm$ 0.18     \\
\multicolumn{1}{l}{} & int2 & 17.44 MB ($\div56.3$)  &  55.03 $\pm$ 1.90 \\ \hline
\end{tabular}
\end{table}

So far FLoCoRA has reduced the number of trainable parameters, which reduces the necessary memory to train the model, and consequently reducing the number of parameters that must be communicated in each round. Next, we investigate the effects of applying an affine quantization scheme~\cite{nagel2021white} for both the client and the server message. We calculate the scaling factor and zero point values per channel for the convolution layers and per column for the FC layer. Normalization layers are not quantized. We use 2/4/8-bit formats to quantify trainable layers. For a model such as ResNet-8 this translates to reductions of 56.3, 32.6 and 17.7 times in the TCC, respectively, for 2/4/8-bits. We included the overhead to transmit the scaling factors and zero points in FP format. Table~\ref{tab:lora_quant} resumes the TTC, as expressed by Equation~\ref{eq2}, to train a model with FLoCoRA and quantization.

\begin{figure}[t]
    \centerline{\includegraphics[width=0.5\textwidth]{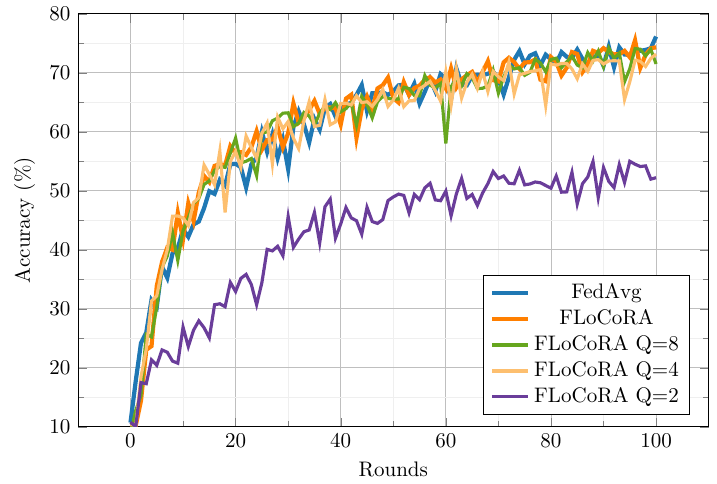}}
    \caption{Convergence behavior between FedAvg, FLoCoRA with rank of 32 and its quantized versions of 2/4/8-bits.}
    \label{fig:lora_quant}
\end{figure}

We show in Figure~\ref{fig:lora_quant} the evolution of the accuracy of the FP FedAvg and FLoCoRA, and also the quantized version of FLoCoRA. The convergence time for FLoCoRA in FP and its quantized version with 8-bits is not affected, demonstrating that it is an effective method for communication savings. As per Table~\ref{tab:lora_quant} the quantized versions suffer further degradation, with 2\% for the 8-bit case. Even with close convergence times, we see that the quantized versions added more instability to the training, this could be a further interesting research direction to improve the quantization scheme with FLoCoRA. We would expect that revisiting the literature on model compression~\cite{nagel2022overcoming,esser2019learned} and combining it with FLoCoRA would find a better quantization scheme. Here, we demonstrate that even a simple technique, based on a round-to-nearest method, is capable of achieving competitive results. 

FedPara~\cite{hyeon2021fedpara} method has been applied to a VGG-16 model, which is compared to FedAvg and a low-rank tensor parameterization based on Tucker decomposition, resulting in communication reductions of 2.8-10.1 times. Base FLoCoRA achieves comparable compression ratios, showing that combining the low-rank adapter for convolution proposed in~\cite{huh2022low} is capable of training a small CNN model from scratch. The quantized version of FLoCoRA further reduces communication cost, resulting in a better trade-off of communication and accuracy.

\begin{table}[h]
\centering
\caption{Comparing LoRA and quantization to ZeroFL and Magnitude Pruning methods.}
\label{tab:lora_vs_all}
\begin{tabular}{ccccc}
\hline
\textbf{Method} & \textbf{Config.} & \textbf{\begin{tabular}[c]{@{}c@{}}Message\\ Size (MB)\end{tabular}} & \textbf{\begin{tabular}[c]{@{}c@{}}TCC \\ (GB)\end{tabular}} & \textbf{Accuracy} \\ \hline
FedAvg & \textbf{Full Model} & 44.7($\div 1.0$) & 62.6 & 84.43 $\pm$  0.36 \\ \hline
\multirow{2}{*}{ZeroFL~\cite{qiu2022zerofl}} & \begin{tabular}[c]{@{}c@{}}90\% SP+\\ 0.2 MR\end{tabular} & 27.3($\div 1.6$) & 38.2 & 81.04 $\pm$ 0.28 \\
 & \begin{tabular}[c]{@{}c@{}}90\% SP+\\ 0.0 MR\end{tabular} & 10.1($\div 4.4$) & 14.1 & 73.87 $\pm$ 0.50 \\ \hline
\multirow{2}{*}{\begin{tabular}[c]{@{}c@{}}Magnitude \\ Pruning~\cite{grativol2023federated}\end{tabular}} & 40\% prune & 27.1 ($\div 1.6$) & 38.0 & 85.20 $\pm$ 0.20 \\
 & 80\% prune & 9.8 ($\div 4.6$) & 13.7 & 80.70 $\pm$ 0.24 \\ \hline
\multirow{6}{*}{FLoCoRA} & r=64 & 9.2 ($\div 4.9$) & 12.9 & 85.17 $\pm$ 0.44 \\
 & r=32 & 4.6 ($\div 9.7$) & 6.5 & 83.90 $\pm$ 0.20 \\
 & r=16 & 2.4 ($\div 18.6$) & 3.3 & 82.33 $\pm$ 0.35 \\
 & r=64, Q=8 & 2.4 ($\div 18.6$) & 3.3 & 85.24 $\pm$ 0.23 \\
 & r=32, Q=8 & 1.2 ($\div 37.3$) & 1.7 & 83.95 $\pm$ 0.32 \\
 & r=16, Q=8 & 0.7 ($\div 63.9$) & 1.0 & 81.89 $\pm$ 1.01 \\ \hline
\end{tabular}
\end{table}

Finally, combining FLoCoRA with the suggested quantization approach, we compare it with the techniques introduced in ZeroFL~\cite{qiu2022zerofl} and with Magnitude Pruning~\cite{grativol2023federated}. Table~\ref{tab:lora_vs_all} illustrates that, although used as a fine-tuning method, LoRA serves as a strong baseline for communication savings in FL. For this experiment, we reproduced the setup proposed by~\cite{qiu2022zerofl,grativol2023federated}, where we have 100 clients, training for 1 local epoch a ResNet-18 model, with an LDA parameter of 1.0, for 700 communication rounds.

 ZeroFL reports their baseline accuracy as 80.62\%, in our work we use the baseline as in Table~\ref{tab:lora_quant}. The difference comes from the batch size used for the clients in the two works. As the client batch size is not reported by ZeroFL, we used the values indicated by Magnitude Pruning. FLoCoRA achieves less accuracy degradation for smaller messages, compared to conventional methods. The same is observed when FLoCoRA is combined with quantization, a reduction between 18.6-63.9 times with 4\% accuracy loss for a case where we spend 0.7 MB per client to train a model with 44.7 MB. Interestingly, compared to our previous experience in Table~\ref{tab:lora_quant}, the quantized versions for ResNet-18 show less degradation than before. We see this to be possible because we have a model that is 9$\times$ larger, being trained 7$\times$ longer, with an easy training scenario. The first results use an LDA of $0.5$ and for ResNet-18 an LDA 1.0, where the higher the LDA parameter, the more identical and close the data distribution between the clients becomes. 

\section{Conclusion}
\label{sec:disc_concl}

In this work, we showed that Low Rank Adapation (LoRA), commonly used for fine-tuning, can be used to train small vision models from scratch in Federated Learning. It allows to decrease the training complexity of previous methods, reducing the amount of communication between server and clients, as well as the training memory requirements and computational power. Moreover, we showed that our method, FLoCoRA, can be deployed with quantization, allowing to further reduce the communication cost, while maintaining strong accuracies. In our setting, using ResNet18 networks on the CIFAR10 dataset, we showed that communication cost is divided by 18.6, with less than 1\% accuracy loss. FLoCoRA paves the way for the deployment of Federated Learning on edge devices, where communication is a bottleneck, and computational power is limited. It also raises some questions to be explored in future work. Indeed, our method is able to train small models without updating the original one. This opens the door to the exploration of model architecture heterogeneity between clients, by using low-rank methods. It would also be possible to explore quantization and rank heterogeneity to further reduce the communication cost.
\bibliographystyle{IEEEtran}
\bibliography{IEEEabrv,main.bib}

\end{document}